\pdfoutput=1
\documentclass[journal,twoside,web]{ieeecolor}
\usepackage{generic}
\usepackage{cite}
\usepackage{amsmath,amssymb,amsfonts}
\usepackage{graphicx}
\usepackage{textcomp}
\usepackage{array}
\usepackage{booktabs}
\usepackage{multirow}
\usepackage[hyphens]{url}
\usepackage{hyperref}
\hypersetup{hidelinks}

\begin{document}
\title{Predicting Live Birth After Fresh Embryo Transfer in IVF/ICSI: Differential ICM/Trophectoderm Morphology Encoding and a Hybrid Attention--Logistic Regression Ensemble}

\author{Abdul Raheem Shahzad$^{\dagger}$, Jinxin Chen$^{\dagger}$, Ming Zhang$^{*}$, Juan Deng$^{*}$
\thanks{This work has been submitted to the IEEE for possible publication. Copyright may be transferred without notice, after which this version may no longer be accessible.}
\thanks{$^{\dagger}$These authors contributed equally to this work and share first authorship.}
\thanks{$^{*}$Corresponding authors: Ming Zhang (whu\_mz@163.com) and Juan Deng (dengjuan@whu.edu.cn).}
\thanks{Abdul Raheem Shahzad and Juan Deng are with the School of Computer Science, Wuhan University, Wuhan, China (e-mail: abdul.raheem.colab@gmail.com; 2024272110002@whu.edu.cn; dengjuan@whu.edu.cn).}
\thanks{Jinxin Chen is with the Department of Obstetrics and Gynecology, Zhongnan Hospital of Wuhan University, Wuhan, China (e-mail: chenjinxin@whu.edu.cn).}
\thanks{Ming Zhang is with the Reproductive Medical Center, Zhongnan Hospital of Wuhan University, Wuhan, Hubei, P.R. China, and also with the Clinical Medicine Research Center of Prenatal Diagnosis and Birth Health in Hubei Province, Wuhan, Hubei, P.R. China (e-mail: whu\_mz@163.com).}}

\maketitle

\begin{abstract}
\textit{Objective:} Predicting live birth before fresh embryo transfer could support shared decision-making in IVF/ICSI, but routine clinical data are small, heterogeneous, and contain pathway-dependent variables prone to outcome leakage. We aimed to construct a curated single-centre IVF/ICSI dataset and develop a prediction model suited to small tabular clinical data. \textit{Materials and Methods:} We assembled 616 fresh embryo-transfer cycles (251 live birth, 365 non-live birth) with clinical, laboratory, and embryological predictors, and embryo morphology encoded with a novel differential weighting of the inner cell mass and trophectoderm. After KNN imputation, robust scaling, and selection of 50 features, we developed a Hybrid Attention Network (HAN), a learnable feature-attention gate followed by a compact multilayer perceptron, and a HAN + logistic-regression (LR) ensemble with a validation-tuned weight. Models were benchmarked against classical machine learning, TabPFN, and text-serialized language models on a held-out test set; AUC-ROC was the primary metric. \textit{Results:} The HAN + LR ensemble achieved the highest test AUC-ROC (0.710; accuracy 0.710, F1 0.630), ahead of the HAN backbone alone (0.700), logistic regression (0.667), the TabPFN + LR hybrid (0.666), and TabPFN (0.650). Text-serialized language models performed near chance (AUC $\le$ 0.61). The differentially weighted trophectoderm feature ranked highest by consensus importance across models. \textit{Discussion:} A compact attention network combined with a linear model outperformed both components and strong foundation-model baselines on a small clinical cohort. \textit{Conclusion:} The curated dataset and HAN + LR ensemble offer a transparent, reproducible basis for IVF/ICSI live-birth prediction; external validation is the next step.
\end{abstract}

\begin{IEEEkeywords}
Blastocyst grading, clinical prediction model, fresh embryo transfer, in vitro fertilization, live birth prediction, machine learning
\end{IEEEkeywords}

\newif\ifshowgraphicalabstract
\showgraphicalabstracttrue
\ifshowgraphicalabstract
\vspace{6pt}
\noindent\includegraphics[width=\columnwidth]{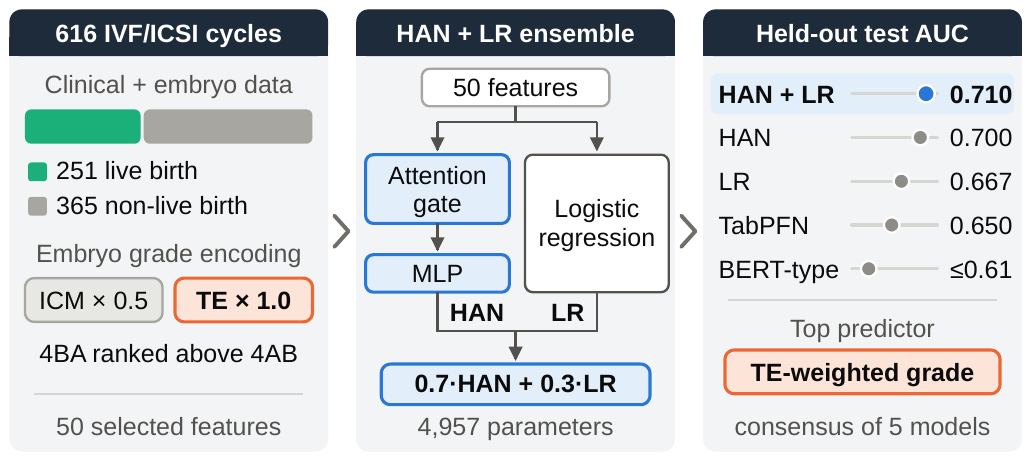}\par
\vspace{3pt}
{\footnotesize\noindent\textbf{Graphical abstract.} Across 616 fresh IVF/ICSI embryo-transfer cycles, a compact hybrid attention network combined with logistic regression (HAN + LR) achieved the highest test AUC-ROC (0.710), outperforming its components, TabPFN, and text-serialized language models. Weighting the trophectoderm above the inner cell mass in blastocyst grading yielded the top-ranked predictor.\par}
\vspace{8pt}
\fi

\section{Background and Significance}\label{sec:background}

\IEEEPARstart{I}{nfertility} affects roughly one in six adults worldwide, and assisted reproductive technology (ART), in particular in vitro fertilization (IVF) and intracytoplasmic sperm injection (ICSI), is now mainstream~\cite{who,hfea}. Despite continued improvement in laboratory and clinical practice, the live-birth rate after a fresh embryo-transfer cycle remains modest, on the order of one in four, and varies substantially between patients~\cite{hfea,zhu}.

Because each cycle carries a considerable physical, emotional, and financial burden, an individualized estimate of live-birth probability before transfer would be clinically valuable, supporting counselling and shared decisions such as the number of embryos to transfer or whether to pursue a freeze-all strategy~\cite{stormlund,eshreEmb,asrmLimits}.

These decisions are made just before transfer, when the clinician already knows the patient's baseline characteristics, stimulation and laboratory results, and embryo morphology. A prediction delivered then, from information available up to that point, is directly actionable. This contrasts with tools estimating cumulative live birth over multiple cycles~\cite{ratna}, which serve a longer-horizon purpose; our work targets the per-transfer decision point, matched to single-cycle structured data.

Many prognostic models have been proposed in ART, but systematic reviews repeatedly identify two limitations: most are not externally validated, and many are at risk of bias under formal appraisal~\cite{henderson,ratna,andaur}. Reporting frameworks such as TRIPOD+AI and PROBAST+AI address these gaps~\cite{tripod,probast}. Much recent methodological energy has gone into image-based deep learning~\cite{mina,theilgaard,louis}, whereas models built on the structured clinical data routinely captured during a cycle remain underexplored, even though such data exist in essentially every clinic.

Tabular clinical data pose their own difficulties: cohorts are small relative to the appetite of modern neural networks, predictors are heterogeneous and partially missing, and several informative variables are pathway-dependent, being themselves consequences of earlier steps in the cycle (for example the number of usable embryos or their grade). Recent advances for tabular learning, including attention-based architectures such as TabNet and FT-Transformer, tabular foundation models such as TabPFN, and serialization approaches such as TabLLM~\cite{tabnet,fttrans,tabpfnIclr,tabllm}, are rarely evaluated head-to-head for IVF live-birth prediction on a single, transparently curated cohort.

This motivates three contributions: (1) a curated single-centre IVF/ICSI dataset for live-birth prediction; (2) a novel numerical encoding of embryo morphology that assigns differential weights to the inner cell mass and trophectoderm of the blastocyst, in place of the equal weighting implicit in conventional Gardner grading; and (3) a prediction model purpose-built for small tabular clinical data and benchmarked against contemporary baselines.

\section{Related Studies}\label{sec:related}

\subsection{Datasets for IVF Outcome Prediction}

Research on IVF outcome prediction draws on two data sources. National registries such as the Human Fertilisation and Embryology Authority (United Kingdom) and SART CORS (United States) offer large samples but capture only a standardized variable set, lacking clinic-level laboratory and embryo detail~\cite{hfea,sartCors,kupka}, whereas single-centre datasets are rich but small, heterogeneous, and affected by missing values. A parallel line of work has released image and time-lapse embryo datasets~\cite{gomez,annotatedBlast,theilgaard}, which target embryo grading rather than cycle-level prediction from structured data. Registry-scale studies have predicted live birth from such databases~\cite{naturalCycle,huang}, and large image corpora have been released for benchmarking~\cite{artDataset,femi}. Our dataset sits in the second group: a curated single-centre collection in which morphology grades are numerically encoded and combined with clinical and laboratory predictors.

\subsection{Traditional Machine-Learning Approaches}

Classical machine learning has been applied repeatedly to IVF/ICSI outcome prediction, where logistic regression remains a strong baseline~\cite{liu,goyal}, and tree-based methods (decision trees, random forests, gradient boosting) have modelled pregnancy and live-birth outcomes~\cite{fu,bai,hassan}. Reported discrimination varies with cohort and outcome definition, large-cohort studies often reporting AUCs of 0.70 to 0.80~\cite{goyal,huang} and single-centre live-birth studies lower and more variable. Broader reviews survey the algorithms used across ART~\cite{ranjini,raef,kim}. Systematic reviews caution that many models use modest samples, are rarely externally validated, and omit calibration~\cite{henderson,ratna,andaur,wu}.

\subsection{Deep Learning, Foundation Models, and Language Models}

Deep learning has entered along two strands. The dominant, image-based strand trains convolutional models on embryo images for grading and selection~\cite{theilgaard,louis,huang,delestro}, including foundation models~\cite{femi} and pipelines localizing informative time points~\cite{dlPipeline,dlEmbryoSel}, with reviews charting laboratory adoption~\cite{riegler,hew,salih} and work on male-factor assessment~\cite{qaderi} and IVF-laboratory automation~\cite{fernandez,siristatidis}. The second strand, to which this work belongs, applies deep and transformer models to tabular data: TabNet and FT-Transformer brought attention to tabular learning~\cite{tabnet,fttrans}, tabular foundation models such as TabPFN give in-context prediction for small datasets~\cite{tabpfnIclr,tabpfnNat}, and language models have been adapted by serializing records into text~\cite{tabllm}. Evidence is mixed: recent benchmarks find well-tuned classical models still match or exceed foundation models on clinical tabular tasks~\cite{benchmark}, underscoring the value of head-to-head evaluation on a single curated cohort.

Two gaps follow. First, structured-data, cycle-level live-birth prediction is under-studied relative to image-based embryo selection. Second, modern tabular model families are rarely benchmarked against one another on the same IVF cohort under a consistent endpoint and transparent pipeline. This study addresses both.

\subsection{Objective}

We set out to (1) construct and characterize a curated dataset of fresh embryo-transfer IVF/ICSI cycles with a binary live-birth label; (2) introduce a differential weighting of the inner cell mass and trophectoderm for numerically encoding blastocyst morphology; and (3) develop and evaluate a Hybrid Attention Network plus logistic-regression (HAN + LR) ensemble for live-birth prediction, benchmarking it against classical machine learning, a tabular foundation model, and text-serialized language models.

\section{Materials and Methods}\label{sec:methods}

\subsection{Data Source and Study Cohort}

We retrospectively assembled clinical data from patients who underwent fresh IVF/ICSI embryo-transfer cycles at the Reproductive Medicine Center of Zhongnan Hospital, Wuhan University, between February 2018 and November 2025. Each treatment cycle constitutes one sample. The curated dataset comprises 616 cycles. The study was approved by the Ethics Committee of Zhongnan Hospital of Wuhan University (Ethical Approval No.~2025346K); given the retrospective design, the requirement for informed consent was waived. Table~\ref{tab::cohort} reports the baseline characteristics of both outcome groups together with the modelling status of every collected variable.

\begin{table*}[!t]
\caption{Baseline Characteristics of the Study Cohort by Outcome Group, With the Modelling Status of Every Collected Variable}
\label{tab::cohort}
\centering
\scriptsize
\setlength{\tabcolsep}{4pt}
\renewcommand{\arraystretch}{1.12}
\begin{tabular}{@{}>{\raggedright\arraybackslash}p{0.35\textwidth}>{\raggedleft\arraybackslash}p{0.13\textwidth}>{\raggedleft\arraybackslash}p{0.13\textwidth}>{\raggedleft\arraybackslash}p{0.07\textwidth}>{\raggedleft\arraybackslash}p{0.055\textwidth}>{\raggedright\arraybackslash}p{0.18\textwidth}@{}}
\toprule
\textbf{Variable} & \textbf{Live birth}\newline (n = 251) & \textbf{Non-live birth}\newline (n = 365) & \textbf{$p$} & \textbf{SMD} & \textbf{Selected} \\
\midrule
\multicolumn{6}{@{}l}{\textbf{Female factors}}\\
Female age (years) & 31.8 $\pm$ 3.7 & 33.1 $\pm$ 4.0 & $<$0.001 & 0.34 & $\checkmark$ \\
Female smoking history, n (\%) &  &  & 0.722 & -- & -- \\
\quad Have & 4 (1.6\%) & 4 (1.1\%) &  &  &  \\
\quad None & 247 (98.4\%) & 361 (98.9\%) &  &  &  \\
Female height (cm) & 160.8 $\pm$ 4.9 & 161.2 $\pm$ 4.5 & 0.269 & 0.09 & $\checkmark$ \\
Female weight (kg) & 59.2 $\pm$ 8.6 & 58.0 $\pm$ 8.0 & 0.062 & 0.15 & -- \textit{(collinear with BMI)} \\
Female BMI (kg/m$^{2}$) & 23.0 $\pm$ 3.2 & 22.3 $\pm$ 2.9 & 0.008 & 0.22 & $\checkmark$ \\
Female basal FSH (mIU/mL) & 7.2 $\pm$ 1.9 & 7.5 $\pm$ 2.2 & 0.103 & 0.13 & $\checkmark$ \\
Female basal LH (mIU/mL) & 6.0 $\pm$ 4.3 & 5.6 $\pm$ 3.5 & 0.228 & 0.10 & $\checkmark$ \\
Female basal E2 (pg/mL) & 38.5 $\pm$ 40.1 & 41.1 $\pm$ 59.4 & 0.528 & 0.05 & -- \\
Female basal PRL (ng/mL) & 18.9 $\pm$ 8.8 & 19.6 $\pm$ 11.2 & 0.380 & 0.07 & -- \\
Female basal T (ng/mL) & 0.3 $\pm$ 1.0 & 0.3 $\pm$ 0.2 & 0.451 & 0.07 & $\checkmark$ \\
Female basal P (ng/mL) & 0.4 $\pm$ 1.3 & 0.4 $\pm$ 1.5 & 0.847 & 0.02 & -- \\
Female AMH (ng/mL) & 3.0 $\pm$ 1.7 & 2.7 $\pm$ 1.6 & 0.035 & 0.18 & $\checkmark$ \\
Female TSH (mIU/L) & 2.1 $\pm$ 1.0 & 2.1 $\pm$ 1.1 & 0.561 & 0.05 & -- \\
Duration of infertility (years) & 2.9 $\pm$ 2.4 & 2.9 $\pm$ 2.2 & 0.759 & 0.03 & -- \\
Female Gn-start-day FSH (mIU/mL) & 6.8 $\pm$ 2.1 & 7.0 $\pm$ 2.4 & 0.402 & 0.07 & $\checkmark$ \\
Female Gn-start-day LH (mIU/mL) & 5.0 $\pm$ 2.6 & 4.9 $\pm$ 2.7 & 0.573 & 0.05 & $\checkmark$ \\
Female Gn-start-day E2 (pg/mL) & 30.7 $\pm$ 15.6 & 33.6 $\pm$ 24.1 & 0.079 & 0.14 & $\checkmark$ \\
Female Gn-start-day P (ng/mL) & 0.3 $\pm$ 0.2 & 0.3 $\pm$ 0.3 & 0.013 & 0.21 & $\checkmark$ \\
Female trigger-day LH (mIU/mL) & 3.1 $\pm$ 2.0 & 3.2 $\pm$ 2.9 & 0.607 & 0.04 & $\checkmark$ \\
Female trigger-day E2 (pg/mL) & 2594.6 $\pm$ 1336.3 & 2433.3 $\pm$ 1388.2 & 0.148 & 0.12 & $\checkmark$ \\
Female trigger-day P (ng/mL) & 0.6 $\pm$ 0.3 & 0.7 $\pm$ 0.3 & 0.322 & 0.08 & -- \\
Trigger-day endometrial thickness (mm) & 11.0 $\pm$ 2.4 & 10.7 $\pm$ 2.4 & 0.126 & 0.13 & $\checkmark$ \\
\addlinespace
\multicolumn{6}{@{}l}{\textbf{Male factors}}\\
Male age (years) & 33.3 $\pm$ 4.4 & 34.5 $\pm$ 4.7 & 0.002 & 0.26 & $\checkmark$ \\
Male smoking history, n (\%) &  &  & 0.883 & -- & -- \\
\quad Have & 91 (36.3\%) & 129 (35.3\%) &  &  &  \\
\quad None & 160 (63.7\%) & 236 (64.7\%) &  &  &  \\
Male height (cm) & 173.4 $\pm$ 5.6 & 173.3 $\pm$ 5.4 & 0.857 & 0.01 & -- \\
Male weight (kg) & 74.7 $\pm$ 12.0 & 75.5 $\pm$ 12.5 & 0.398 & 0.07 & $\checkmark$ \\
Male BMI (kg/m$^{2}$) & 24.8 $\pm$ 3.4 & 25.1 $\pm$ 3.7 & 0.295 & 0.09 & $\checkmark$ \\
Male serum FSH (mIU/mL) & 4.8 $\pm$ 4.1 & 4.8 $\pm$ 3.6 & 0.891 & 0.01 & -- \\
Male serum LH (mIU/mL) & 4.6 $\pm$ 2.9 & 4.7 $\pm$ 2.9 & 0.514 & 0.05 & $\checkmark$ \\
Male serum PRL (ng/mL) & 13.1 $\pm$ 14.7 & 13.2 $\pm$ 7.5 & 0.923 & 0.01 & $\checkmark$ \\
Male serum T (ng/mL) & 4.5 $\pm$ 2.9 & 5.3 $\pm$ 16.7 & 0.362 & 0.07 & -- \\
Sperm concentration (10$^{6}$/mL) & 45.5 $\pm$ 36.5 & 45.7 $\pm$ 33.6 & 0.954 & 0.00 & -- \\
Progressive motility (PR, \%) & 37.3 $\pm$ 18.0 & 39.0 $\pm$ 18.1 & 0.274 & 0.09 & $\checkmark$ \\
Non-progressive motility (NP, \%) & 4.3 $\pm$ 4.2 & 4.5 $\pm$ 6.7 & 0.716 & 0.03 & -- \\
Immotility (IM, \%) & 57.6 $\pm$ 19.6 & 56.4 $\pm$ 19.4 & 0.445 & 0.06 & $\checkmark$ \\
Normal sperm morphology (\%) & 6.6 $\pm$ 5.8 & 6.2 $\pm$ 4.3 & 0.414 & 0.07 & -- \\
Sperm DFI (\%) & 11.3 $\pm$ 5.6 & 12.1 $\pm$ 5.8 & 0.188 & 0.13 & $\checkmark$ \\
\bottomrule
\end{tabular}
\par\vspace{3pt}
\begin{minipage}{\textwidth}\scriptsize\raggedright
Continuous variables are mean $\pm$ SD; categorical variables are n (\%). $p$ values are from Welch $t$ tests for continuous variables and $\chi^2$ or Fisher exact tests for categorical variables, reported without correction for multiple comparisons; SMD is the absolute standardised mean difference. The final column marks variables retained by feature selection ($\checkmark$). Selection operated on encoded features, in which each category of a categorical variable becomes a separate binary feature, so a variable is marked if any of its categories was selected. The 50 selected features therefore correspond to 47 variables, because three categorical variables (ovarian stimulation protocol, method of fertilization, and number of cleavage-stage embryos transferred) each had two categories selected. Variables excluded for a documented reason are annotated.
\end{minipage}
\end{table*}

\begin{table*}[!t]
\addtocounter{table}{-1}
\renewcommand{\theHtable}{\arabic{table}.continued}
\caption{(Continued)}
\centering
\scriptsize
\setlength{\tabcolsep}{4pt}
\renewcommand{\arraystretch}{1.12}
\begin{tabular}{@{}>{\raggedright\arraybackslash}p{0.35\textwidth}>{\raggedleft\arraybackslash}p{0.13\textwidth}>{\raggedleft\arraybackslash}p{0.13\textwidth}>{\raggedleft\arraybackslash}p{0.07\textwidth}>{\raggedleft\arraybackslash}p{0.055\textwidth}>{\raggedright\arraybackslash}p{0.18\textwidth}@{}}
\toprule
\textbf{Variable} & \textbf{Live birth}\newline (n = 251) & \textbf{Non-live birth}\newline (n = 365) & \textbf{$p$} & \textbf{SMD} & \textbf{Selected} \\
\midrule
\multicolumn{6}{@{}l}{\textbf{Stimulation and laboratory}}\\
Ovarian stimulation protocol, n (\%) &  &  & 0.795 & -- & $\checkmark$ \\
\quad Antagonist regimen & 199 (79.3\%) & 296 (81.1\%) &  &  &  \\
\quad Extra Long Plan & 3 (1.2\%) & 7 (1.9\%) &  &  &  \\
\quad Long protocol for early follicular phase & 21 (8.4\%) & 29 (7.9\%) &  &  &  \\
\quad Luteal phase extension protocol & 26 (10.4\%) & 29 (7.9\%) &  &  &  \\
\quad Microstimulation Program & 2 (0.8\%) & 4 (1.1\%) &  &  &  \\
Method of fertilization, n (\%) &  &  & 0.315 & -- & $\checkmark$ \\
\quad ICSI & 20 (8.0\%) & 20 (5.5\%) &  &  &  \\
\quad IVF & 224 (89.2\%) & 338 (92.6\%) &  &  &  \\
\quad IVF+RICSI & 3 (1.2\%) & 1 (0.3\%) &  &  &  \\
\quad RICSI & 4 (1.6\%) & 6 (1.6\%) &  &  &  \\
Duration of Gn stimulation (days) & 10.0 $\pm$ 1.6 & 9.7 $\pm$ 1.6 & 0.045 & 0.17 & $\checkmark$ \\
Total Gn dose (IU) & 2522.0 $\pm$ 736.7 & 2527.2 $\pm$ 667.3 & 0.928 & 0.01 & $\checkmark$ \\
Total FSH dose (IU) & 2102.0 $\pm$ 755.1 & 2158.3 $\pm$ 606.4 & 0.326 & 0.08 & $\checkmark$ \\
Total HMG dose (IU) & 374.0 $\pm$ 514.2 & 363.1 $\pm$ 433.1 & 0.784 & 0.02 & -- \\
Total LH dose (IU) & 43.0 $\pm$ 113.8 & 23.8 $\pm$ 97.7 & 0.030 & 0.18 & $\checkmark$ \\
Urinary hCG trigger dose (IU) & 6374.5 $\pm$ 4797.0 & 6608.2 $\pm$ 5128.3 & 0.564 & 0.05 & -- \\
Recombinant hCG trigger dose ($\mu$g) & 12.9 $\pm$ 55.5 & 7.5 $\pm$ 42.8 & 0.194 & 0.11 & $\checkmark$ \\
GnRH agonist trigger dose (mg) & 0.1 $\pm$ 0.1 & 0.1 $\pm$ 0.1 & 0.633 & 0.04 & -- \\
Trigger type, n (\%) &  &  & 0.704 & -- & -- \\
\quad Double trigger & 106 (42.2\%) & 161 (44.1\%) &  &  &  \\
\quad Single trigger & 145 (57.8\%) & 204 (55.9\%) &  &  &  \\
\addlinespace
\multicolumn{6}{@{}l}{\textbf{Oocyte retrieval and embryo development}}\\
Number of follicles punctured & 12.6 $\pm$ 5.2 & 11.6 $\pm$ 5.5 & 0.019 & 0.19 & $\checkmark$ \\
Number of oocytes retrieved & 10.2 $\pm$ 4.3 & 9.2 $\pm$ 4.5 & 0.003 & 0.24 & $\checkmark$ \\
Number of MII oocytes & 9.9 $\pm$ 4.2 & 8.9 $\pm$ 4.5 & 0.006 & 0.23 & $\checkmark$ \\
Number of 2PN zygotes & 7.2 $\pm$ 3.4 & 6.1 $\pm$ 3.3 & $<$0.001 & 0.31 & $\checkmark$ \\
Number of multipronuclear zygotes & 0.0 $\pm$ 0.0 & 0.0 $\pm$ 0.0 & -- & -- & -- \\
Number of high-quality embryos & 4.9 $\pm$ 2.7 & 4.2 $\pm$ 2.6 & 0.002 & 0.26 & $\checkmark$ \\
Number of embryos available at D3 & 6.6 $\pm$ 3.2 & 5.7 $\pm$ 3.1 & $<$0.001 & 0.30 & $\checkmark$ \\
Number of blastocyst-cultured embryos & 6.5 $\pm$ 3.2 & 5.7 $\pm$ 3.2 & 0.002 & 0.27 & $\checkmark$ \\
Number of blastocysts formed & 5.2 $\pm$ 2.8 & 4.6 $\pm$ 2.9 & 0.006 & 0.24 & $\checkmark$ \\
Types of embryos transferred, n (\%) &  &  & 0.001 & -- & $\checkmark$ \\
\quad Blastocyst stage embryo & 172 (68.5\%) & 201 (55.1\%) &  &  &  \\
\quad Cleavage stage embryo & 79 (31.5\%) & 164 (44.9\%) &  &  &  \\
Embryo morphology (52 distinct values) & -- & -- & 0.009 & -- & -- \textit{(encoded as 10 columns)} \\
Number of cleavage-stage embryos transferred & 1.5 $\pm$ 0.5 & 1.3 $\pm$ 0.4 & $<$0.001 & 0.56 & $\checkmark$ \\
Number of blastocysts transferred & 1.0 $\pm$ 0.2 & 1.0 $\pm$ 0.2 & 0.420 & 0.08 & $\checkmark$ \\
Blastocyst 1 expansion quality & 1.4 $\pm$ 0.9 & 1.1 $\pm$ 1.0 & $<$0.001 & 0.28 & $\checkmark$ \\
Blastocyst 1 ICM (weighted) & 0.8 $\pm$ 0.6 & 0.7 $\pm$ 0.6 & 0.002 & 0.25 & $\checkmark$ \\
Blastocyst 1 TE (weighted) & 1.3 $\pm$ 1.0 & 1.0 $\pm$ 1.0 & $<$0.001 & 0.32 & $\checkmark$ \\
Blastocyst 2 expansion quality & 0.0 $\pm$ 0.3 & 0.0 $\pm$ 0.3 & 0.775 & 0.02 & -- \\
Blastocyst 2 ICM (weighted) & 0.0 $\pm$ 0.1 & 0.0 $\pm$ 0.2 & 0.626 & 0.04 & -- \\
Blastocyst 2 TE (weighted) & 0.0 $\pm$ 0.2 & 0.0 $\pm$ 0.3 & 0.575 & 0.05 & -- \\
Cleavage embryo 1 grade & 0.9 $\pm$ 1.4 & 1.3 $\pm$ 1.5 & 0.001 & 0.26 & $\checkmark$ \\
Cleavage 1 cell-count quality & 0.8 $\pm$ 1.2 & 1.1 $\pm$ 1.3 & 0.002 & 0.26 & $\checkmark$ \\
Cleavage embryo 2 grade & 0.5 $\pm$ 1.1 & 0.3 $\pm$ 0.9 & 0.059 & 0.16 & $\checkmark$ \\
Cleavage 2 cell-count quality & 0.4 $\pm$ 1.0 & 0.3 $\pm$ 0.8 & 0.084 & 0.14 & $\checkmark$ \\
\bottomrule
\end{tabular}
\end{table*}

\subsection{Outcome Definition}

The prediction target was live birth, defined as the delivery of one or more living infants after at least 24 weeks of gestation, in line with standard fertility-care glossaries~\cite{zegers}, versus non-live birth; the non-live birth group comprised cycles with no pregnancy (n = 293), a biochemical pregnancy (n = 27), or a miscarriage (n = 45). Cycles were accordingly assigned to the live birth group or the non-live birth group, and this binary outcome served as the label for all models.

\subsection{Predictors and Embryo-Morphology Encoding}

Candidate predictors were organized into the four clinical domains used throughout this paper and in Table~\ref{tab::cohort}. Female factors comprised age, height, weight, body-mass index, smoking history, duration of infertility, the basal, Gn-start-day and trigger-day hormonal profile (follicle-stimulating and luteinizing hormone, estradiol, progesterone, prolactin, testosterone, anti-Mullerian and thyroid-stimulating hormone), and trigger-day endometrial thickness. Male factors comprised age, height, weight, body-mass index, smoking history, serum follicle-stimulating and luteinizing hormone, prolactin and testosterone, and semen parameters (concentration, progressive and non-progressive motility, immotility, normal morphology, DNA fragmentation index). Stimulation and laboratory variables comprised the ovarian-stimulation protocol, broadly consistent with guideline practice~\cite{eshreStim}, the method of fertilization, the duration of gonadotropin stimulation, the total gonadotropin, follicle-stimulating hormone, menopausal gonadotropin and luteinizing hormone doses, and the trigger type and dose. Oocyte-retrieval and embryo-development variables comprised the numbers of punctured follicles, retrieved and MII oocytes, 2PN and multipronuclear zygotes, high-quality embryos, embryos available at day 3, blastocyst-cultured embryos and blastocysts formed, with the type and number of embryos transferred and the encoded morphology. Embryo morphology, recorded as qualitative grades, was converted into numerical features using the Peter system for cleavage-stage embryos and the Gardner system for blastocysts, so that morphology could be used directly irrespective of transfer stage, an issue of ongoing debate~\cite{glujovsky}.

\subsection{A Differential ICM/TE Weighting for Blastocyst Grading}\label{sec:encoding}

\begin{figure}[!t]
    \centering
    \IfFileExists{figures/fig_encoding_scheme.pdf}{\includegraphics[width=\columnwidth]{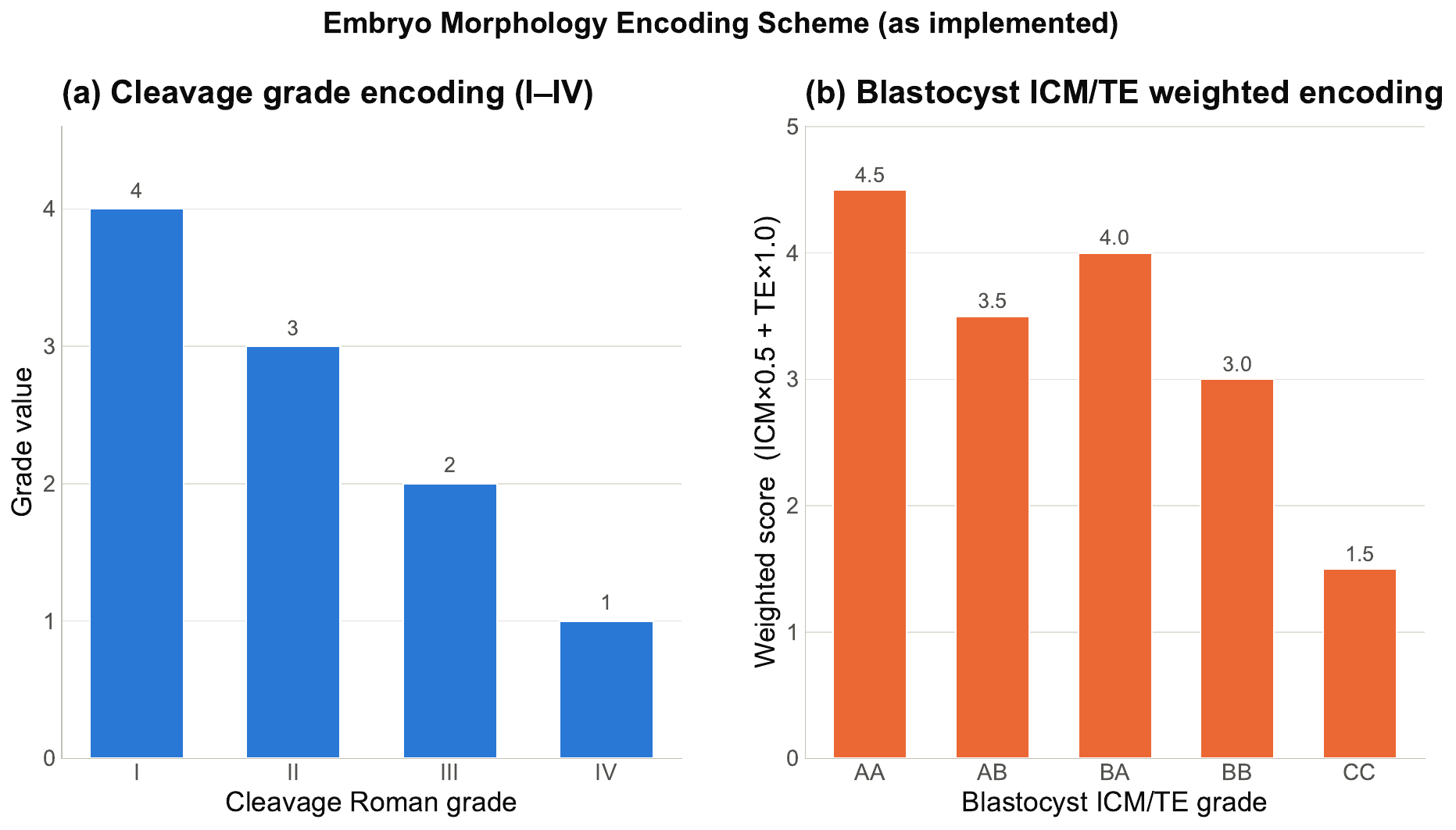}}{\fbox{\begin{minipage}[c][3.6cm][c]{0.86\columnwidth}\centering\textbf{Figure} --- add \texttt{figures/fig\_encoding\_scheme.pdf}\par\smallskip\footnotesize embryo morphology encoding scheme\end{minipage}}}
    \caption{Diagram of the embryo morphology encoding scheme.}
    \label{fig::encoding}
\end{figure}

A specific methodological contribution of this work concerns how embryo morphology is encoded. The raw morphology text, describing one or two transferred embryos in either the Gardner blastocyst system (expansion, inner cell mass [ICM], trophectoderm [TE]) or the cleavage-stage system (grade and cell number), was converted into ten numeric columns, with separate slots for up to two embryos and for the two systems. Blastocyst ICM and TE letters were mapped A/B/C to 3/2/1 and cleavage Roman grades I-IV to 4-1. Rather than treat ICM and TE as equally informative, as conventional Gardner grading implicitly does, we weight them differentially: informed by a correlation analysis in this cohort, in which TE was more strongly associated with live birth than ICM, the composite stores ICM at half weight and TE at full weight (ICM $\times$ 0.5, TE $\times$ 1.0). This resolves the ambiguity between blastocysts sharing an expansion stage but with mirror-image grades, ranking the TE-superior 4BA above 4AB. Two further non-monotonic facts are captured by inverted-U optimality scores: cleavage cell count (best around eight cells) and blastocyst expansion (best at intermediate stages). The weighting is applied during dataset construction, giving models a more expressive morphology feature than an equal-weight grade sum; the TE-weighted feature was among the 50 selected and, in cross-checks, among the most influential. Fig.~\ref{fig::encoding} summarizes the encoding scheme.

\subsection{Preprocessing and Feature Selection}

Records were cleaned to coerce stray text entries to missing values. To prevent leakage, all fitting was confined to the training split: numerical variables were imputed with a k-nearest-neighbour imputer ($k = 5$) and scaled with a robust scaler (median and interquartile range). Female weight was dropped as collinear with body-mass index, and categorical variables were one-hot encoded with the first level as reference. Of the 71 collected predictors listed in Table~\ref{tab::cohort}, two were removed before modelling: the raw embryo-morphology text, which is represented by the ten numeric columns described above, and female weight. The remaining 69 variables comprised 61 numerical variables (including those ten embryo columns) and 8 categorical variables, and one-hot encoding the latter produced 15 dummies, giving 76 candidate features. From these, the 50 most informative were retained using univariate selection (SelectKBest with the ANOVA F-statistic), fit on the training split only, a deliberately simple, transparent filter suited to a small cohort. The same preprocessed feature set and the same stratified partition were used for every model, so that performance differences reflect the models rather than differences in data handling. Because selection operates on encoded features, a categorical variable can contribute more than one: three variables (ovarian stimulation protocol, method of fertilization, and number of cleavage-stage embryos transferred) each had two categories selected, so the 50 selected features correspond to 47 of the 71 collected variables; Table~\ref{tab::cohort} records the modelling status of every collected variable alongside its distribution in each outcome group.

\subsection{Model: Hybrid Attention Network and the HAN + LR Ensemble}\label{sec:model}

The HAN backbone has two stages. A Feature Attention Module first applies a global, dataset-level learnable importance gate: a weight vector is passed through a sigmoid and multiplied element-wise with the input features, allowing the network to up- or down-weight predictors. Unlike per-sample attention in large transformers, this gate is shared across the cohort, keeping the parameter count low and suiting a small clinical dataset. The gated features then pass through a compact multilayer perceptron (50$\rightarrow$24 with layer normalization, ReLU, and dropout 0.25; 24$\rightarrow$12 with ReLU and dropout 0.15; 12$\rightarrow$1 with a sigmoid), yielding a live-birth probability. Each HAN instance contains approximately 1,635 trainable parameters. To reduce variance from random initialization, the backbone is realized as an ensemble of three independently seeded instances whose probabilities are averaged.

The proposed model combines the HAN backbone with an L2-regularized logistic-regression (LR) model trained on the same features. The final probability is a convex combination $p = w \cdot p_{\mathrm{HAN}} + (1-w) \cdot p_{\mathrm{LR}}$, with the mixing weight selected by a grid search that maximized validation AUC. The search selected $w = 0.70$, weighting the attention backbone more heavily than the linear model, reflecting that the HAN backbone was the stronger of the two components on this cohort while logistic regression still contributed complementary structure. Counting all components, the ensemble comprises 4,957 parameters (three HAN instances, the LR model, and the mixing weight).

\subsection{Training Configuration}

HAN instances were trained by minimizing binary cross-entropy with the AdamW optimizer (learning rate 0.002, weight decay 0.001), for up to 80 epochs with a batch size of 32, using Xavier initialization, gradient clipping at a maximum norm of 1.0, and early stopping on the best validation AUC. Key settings are summarized in Table~\ref{tab::config}.

\begin{table}[!t]
\caption{HAN\,+\,LR Architecture and Training Configuration}
\label{tab::config}
\centering
\footnotesize
\begin{tabular}{@{}>{\raggedright\arraybackslash}p{0.30\columnwidth}>{\raggedright\arraybackslash}p{0.63\columnwidth}@{}}
\toprule
\textbf{Component} & \textbf{Setting} \\
\midrule
Feature Attention Module & Learnable per-feature gate, sigmoid, element-wise multiplicative \\
MLP layers & $50\rightarrow24$ (LayerNorm, ReLU, dropout 0.25); $24\rightarrow12$ (ReLU, dropout 0.15); $12\rightarrow1$ (sigmoid) \\
HAN ensemble & 3 seeds (42, 123, 456), probabilities averaged \\
Parameters & $\approx$1,635 per HAN instance; 4,957 total \\
Ensemble rule & $p = w\,p_{\mathrm{HAN}} + (1-w)\,p_{\mathrm{LR}}$; $w=0.70$ by validation grid search \\
Optimizer / loss & AdamW (lr 0.002, weight decay 0.001); binary cross-entropy \\
Schedule & $\le$80 epochs, batch 32, early stopping on validation AUC \\
Stability & Xavier initialization; gradient clipping at max-norm 1.0 \\
\bottomrule
\end{tabular}
\end{table}

\subsection{Comparator Models}

We compared the HAN + LR ensemble with: classical machine-learning models (logistic regression, random forest, gradient boosting, support-vector machine, k-nearest neighbours, naive Bayes, and XGBoost); a tabular foundation model (TabPFN) and a TabPFN + LR hybrid; and a family of pre-trained language models adapted to tabular data by text serialization, in which each record is rendered as a natural-language string of feature–value pairs and embedded (MiniLM, DistilBERT, BERT, ELECTRA, and RoBERTa)~\cite{tabpfnIclr,tabllm}. Classical models used standard, lightly tuned configurations; the TabPFN + LR hybrid used the same validation-tuned weighting as the proposed ensemble, so that only the neural component differs. The language-model pipeline used one fixed serialization template across all encoders.

\subsection{Evaluation}

Data were partitioned by stratified sampling into 523 training and 93 held-out test cycles (approximately 85\% and 15\%), preserving the class ratio. Cross-validation within the training set was used for model selection and tuning, including the ensemble mixing weight; final performance was reported on the held-out test set. The primary metric was the area under the receiver-operating-characteristic curve (AUC-ROC), which is threshold-independent and appropriate for a moderately imbalanced binary outcome; accuracy, precision, recall, and F1 were reported as secondary metrics at the default 0.5 decision threshold. Bootstrap resampling of the test set was used to characterize the stability of the estimates. Reporting follows the TRIPOD+AI checklist~\cite{tripod}.

\section{Results}\label{sec:results}

\subsection{Cohort}

The curated cohort comprised 616 fresh embryo-transfer cycles, partitioned into 523 training and 93 test cycles (of which 38 were live births). The live birth group was the minority (251 cycles, 40.7\%) and the non-live birth group the majority (365 cycles, 59.3\%); we preserved this ratio by stratified sampling rather than applying explicit resampling. Baseline characteristics by outcome group are given in Table~\ref{tab::cohort}.

The 50 selected features spanned all four predictor domains (Table~\ref{tab::cohort}) but were dominated by oocyte-retrieval and embryo-development variables and by the female hormonal profile. This composition is clinically coherent, since embryo quality and yield are among the strongest correlates of transfer success, but it also signals the pathway-dependence discussed below.

\subsection{Primary Model Comparison}

\begin{figure}[!t]
    \centering
    \IfFileExists{figures/fig_roc_curves.pdf}{\includegraphics[width=\columnwidth]{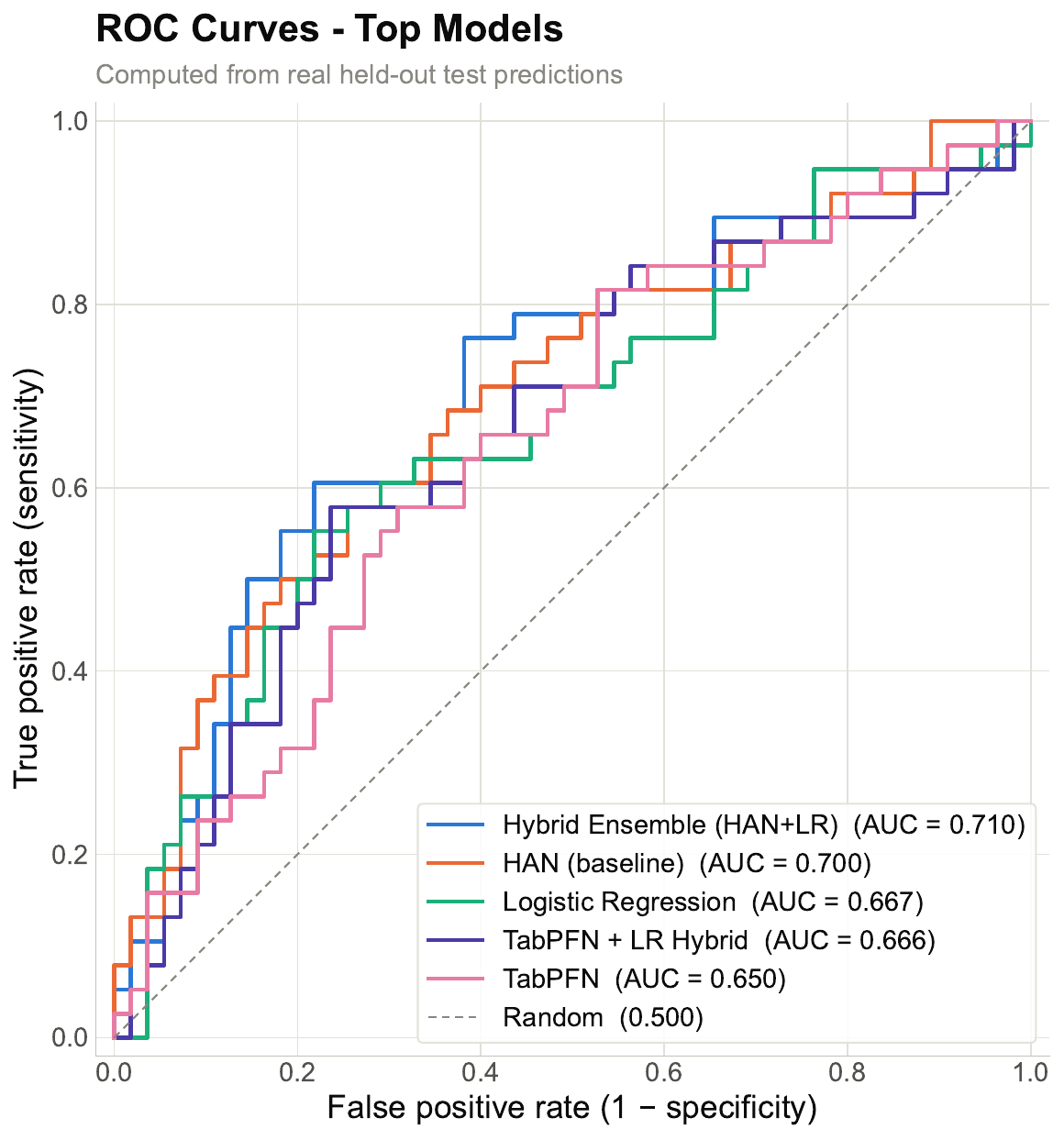}}{\fbox{\begin{minipage}[c][3.6cm][c]{0.86\columnwidth}\centering\textbf{Figure} --- add \texttt{figures/fig\_roc\_curves.pdf}\par\smallskip\footnotesize AUC-ROC curves\end{minipage}}}
    \caption{Receiver-operating-characteristic (AUC-ROC) curves for the evaluated models on the held-out test set.}
    \label{fig::roc}
\end{figure}

Across the eleven models evaluated on the held-out test set, the HAN + LR ensemble achieved the highest AUC-ROC (0.710), with an accuracy of 0.710 and F1 of 0.630 (Table~\ref{tab::results}). It outperformed the HAN backbone evaluated alone (0.700), the strongest classical baseline logistic regression (0.667), the TabPFN + LR hybrid (0.666), and standalone TabPFN (0.650). The remaining models (support-vector machine, gradient boosting, XGBoost, k-nearest neighbours, naive Bayes, and random forest) trailed with AUC-ROC between 0.56 and 0.63; receiver-operating-characteristic curves for the models are shown in Fig.~\ref{fig::roc}.

\begin{table}[!t]
\caption{Test-Set Performance of the Eleven Evaluated Models, Ranked by AUC-ROC}
\label{tab::results}
\centering
\resizebox{\columnwidth}{!}{%
\begin{tabular}{lccccc}
\toprule
\textbf{Model} & \textbf{AUC} & \textbf{Acc} & \textbf{Prec} & \textbf{Rec} & \textbf{F1} \\
\midrule
\textbf{HAN\,+\,LR (proposed)} & \textbf{0.710} & 0.710 & 0.657 & 0.605 & 0.630 \\
HAN backbone (alone) & 0.700 & 0.667 & 0.590 & 0.605 & 0.597 \\
Logistic regression & 0.667 & 0.645 & 0.558 & 0.632 & 0.593 \\
TabPFN\,+\,LR & 0.666 & 0.656 & 0.615 & 0.421 & 0.500 \\
TabPFN & 0.650 & 0.613 & 0.545 & 0.316 & 0.400 \\
Support-vector machine & 0.632 & 0.624 & 0.588 & 0.263 & 0.364 \\
Gradient boosting & 0.627 & 0.591 & 0.500 & 0.395 & 0.441 \\
XGBoost & 0.624 & 0.613 & 0.533 & 0.421 & 0.471 \\
$k$-nearest neighbours & 0.613 & 0.645 & 0.586 & 0.447 & 0.507 \\
Na\"ive Bayes & 0.581 & 0.570 & 0.471 & 0.421 & 0.444 \\
Random forest & 0.556 & 0.591 & 0.500 & 0.263 & 0.345 \\
\bottomrule
\end{tabular}}
\par\vspace{3pt}
\begin{minipage}{\columnwidth}\scriptsize\raggedright Higher is better for all metrics.\end{minipage}
\end{table}

Two observations are notable. First, the HAN + LR ensemble improved on its own components by +0.010 AUC over the HAN backbone (0.700) and +0.043 over logistic regression (0.667); the gain is modest but consistent, and the HAN backbone alone again outperformed logistic regression and every other single model, ranking second only to the hybrid. Second, unlike the HAN backbone, the tabular foundation model did not help the linear model: standalone TabPFN reached only 0.650 and the TabPFN + LR hybrid 0.666, neither exceeding logistic regression (0.667), so the complementary signal came specifically from pairing the linear model with the attention backbone rather than a foundation model.

The threshold-dependent secondary metrics highlight different operating behaviours that AUC-ROC alone conceals. TabPFN, the support-vector machine, and random forest favoured precision at the cost of recall (0.263 to 0.316), flagging few cycles as live births, whereas logistic regression achieved the highest recall (0.632) at lower precision (0.558). The HAN + LR ensemble led on four of the five metrics, AUC-ROC, accuracy (0.710), precision (0.657), and F1 (0.630), while maintaining a recall of 0.605. Because these metrics use a fixed 0.5 threshold, they are sensitive to the operating point and to sampling variation; AUC-ROC is therefore the more reliable basis for ranking, and the bootstrap intervals around them were wide.

\subsection{Feature Importance Across Models}

\begin{figure}[!t]
    \centering
    \IfFileExists{figures/fig_feature_importance.pdf}{\includegraphics[width=\columnwidth]{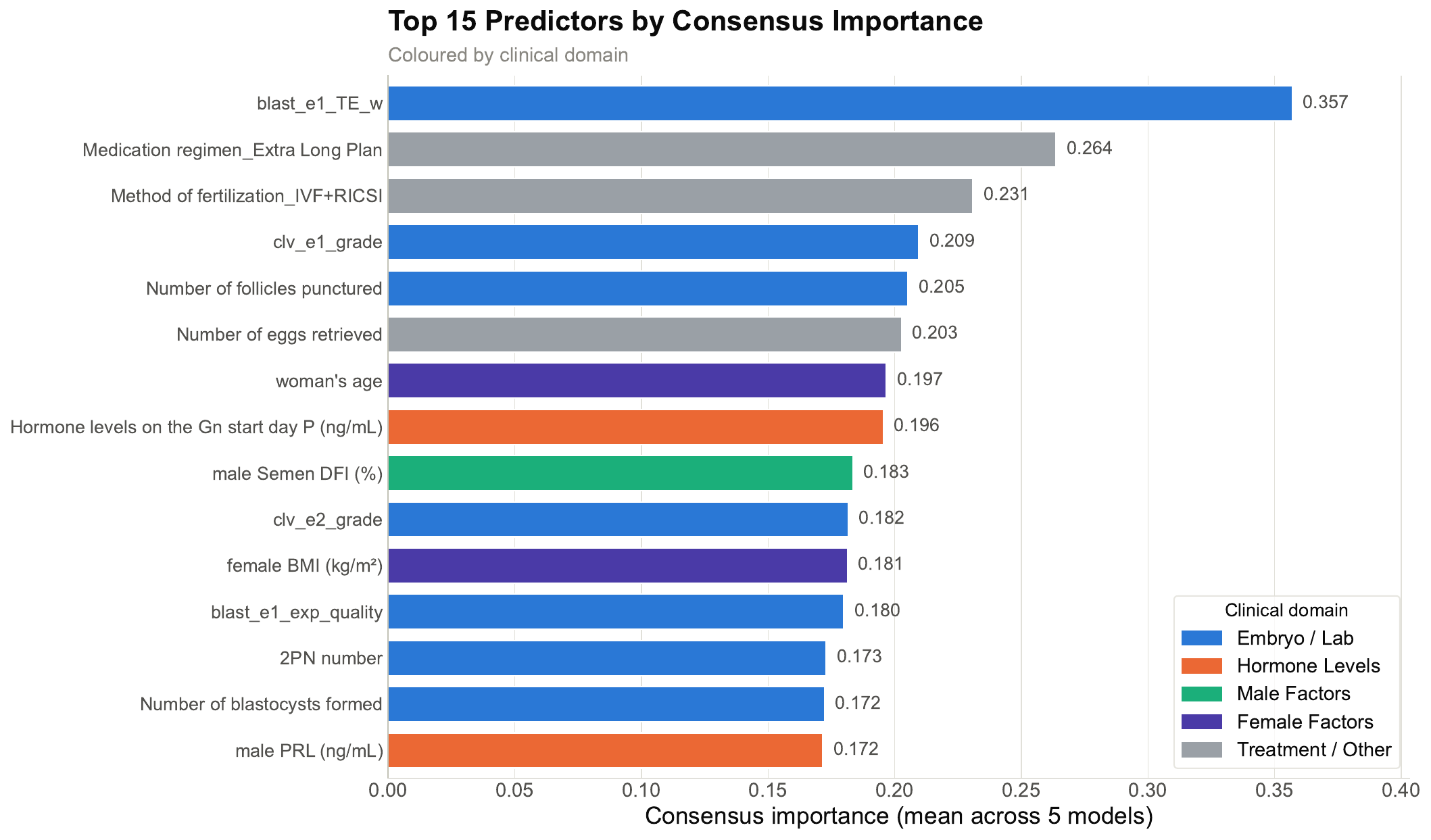}}{\fbox{\begin{minipage}[c][3.6cm][c]{0.86\columnwidth}\centering\textbf{Figure} --- add \texttt{figures/fig\_feature\_importance.pdf}\par\smallskip\footnotesize consensus feature importance\end{minipage}}}
    \caption{Top 15 predictors by consensus importance, averaged across the five models that expose an importance measure and coloured by the clinical domains defined in Table~\ref{tab::cohort}.}
    \label{fig::importance}
\end{figure}

\begin{figure}[!t]
    \centering
    \IfFileExists{figures/fig_importance_by_model.pdf}{\includegraphics[width=\columnwidth]{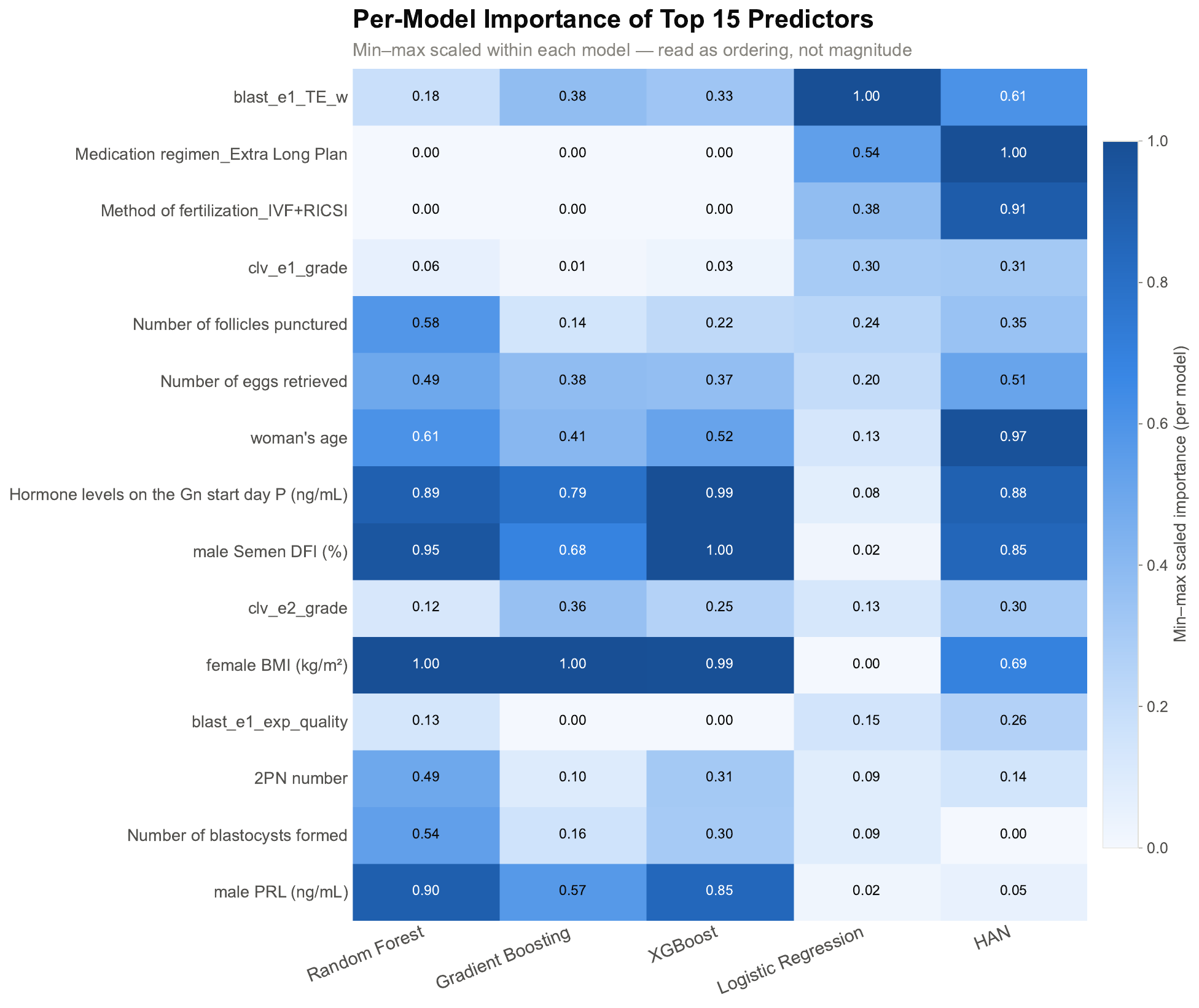}}{\fbox{\begin{minipage}[c][3.6cm][c]{0.86\columnwidth}\centering\textbf{Figure} --- add \texttt{figures/fig\_importance\_by\_model.pdf}\par\smallskip\footnotesize per-model importance heatmap\end{minipage}}}
    \caption{Relative importance of each of the top 15 predictors within each model, min-max scaled per model so that differently scaled measures can be compared. Scaling amplifies the narrow spread of the HAN attention weights and should be read as ordering, not magnitude.}
    \label{fig::bymodel}
\end{figure}

To characterize which predictors each model relied on, we extracted each model's native importance measure: Gini importance (random forest), Friedman MSE improvement (gradient boosting), gain-based weight (XGBoost), absolute coefficient magnitude (logistic regression), and the learned attention weights (HAN). Because these live on different scales, we min-max scaled them within each model and averaged the normalized values into a consensus ranking (Table~\ref{tab::importance}, Figs.~\ref{fig::importance} and~\ref{fig::bymodel}).

\begin{table*}[!t]
\caption{Top 15 Predictors by Consensus Importance, With the Native Importance Reported by Each Model}
\label{tab::importance}
\centering
\footnotesize
\setlength{\tabcolsep}{4pt}
\begin{tabular}{@{}>{\raggedright\arraybackslash}p{0.235\textwidth}>{\raggedright\arraybackslash}p{0.215\textwidth}>{\raggedleft\arraybackslash}p{0.085\textwidth}rrrrr@{}}
\toprule
\textbf{Feature} & \textbf{Domain} & \textbf{Consensus (\%)} & \textbf{RF} & \textbf{GB} & \textbf{XGB} & \textbf{LR} & \textbf{HAN} \\
\midrule
Blastocyst 1 TE (weighted) & Oocyte retrieval and embryo development & 4.53 & 0.008 & 0.019 & 0.014 & 1.208 & 0.537 \\
Ovarian stimulation protocol (Extra Long Plan) & Stimulation and laboratory & 3.35 & 0.001 & 0.000 & 0.000 & 0.759 & 0.560 \\
Method of fertilization (IVF+RICSI) & Stimulation and laboratory & 2.93 & 0.001 & 0.000 & 0.000 & 0.599 & 0.555 \\
Cleavage embryo 1 grade & Oocyte retrieval and embryo development & 2.66 & 0.003 & 0.000 & 0.001 & 0.523 & 0.520 \\
Number of follicles punctured & Oocyte retrieval and embryo development & 2.61 & 0.024 & 0.007 & 0.009 & 0.464 & 0.522 \\
Number of oocytes retrieved & Oocyte retrieval and embryo development & 2.57 & 0.020 & 0.019 & 0.015 & 0.427 & 0.532 \\
Female age (years) & Female factors & 2.50 & 0.025 & 0.020 & 0.022 & 0.359 & 0.558 \\
Female Gn-start-day P (ng/mL) & Female factors & 2.48 & 0.036 & 0.040 & 0.041 & 0.308 & 0.553 \\
Sperm DFI (\%) & Male factors & 2.33 & 0.038 & 0.034 & 0.041 & 0.252 & 0.551 \\
Cleavage embryo 2 grade & Oocyte retrieval and embryo development & 2.31 & 0.005 & 0.018 & 0.011 & 0.355 & 0.520 \\
Female BMI (kg/m$^2$) & Female factors & 2.30 & 0.040 & 0.050 & 0.041 & 0.233 & 0.542 \\
Blastocyst 1 expansion quality & Oocyte retrieval and embryo development & 2.28 & 0.006 & 0.000 & 0.000 & 0.377 & 0.517 \\
Number of 2PN zygotes & Oocyte retrieval and embryo development & 2.19 & 0.020 & 0.005 & 0.013 & 0.316 & 0.510 \\
Number of blastocysts formed & Oocyte retrieval and embryo development & 2.19 & 0.022 & 0.008 & 0.013 & 0.317 & 0.502 \\
Male serum PRL (ng/mL) & Male factors & 2.18 & 0.036 & 0.029 & 0.035 & 0.253 & 0.505 \\
\bottomrule
\end{tabular}
\par\vspace{3pt}
\begin{minipage}{\textwidth}\scriptsize\raggedright Variable names and clinical domains follow Table~\ref{tab::cohort}. Values are Gini importance (random forest), Friedman MSE improvement (gradient boosting), gain-based weight (XGBoost), absolute coefficient magnitude (logistic regression), and the learned attention weight (HAN). Note the differing native scales, and in particular the narrow range of the HAN attention weights.\end{minipage}
\end{table*}

Three patterns emerged. First, the differentially weighted blastocyst trophectoderm feature (Blastocyst 1 TE) was the highest-ranked predictor overall (4.53\% of consensus importance) and carried by far the largest logistic-regression coefficient, which supports the encoding scheme introduced above; the cleavage embryo grades and the blastocyst expansion-quality score also appeared among the top 15, with the weighted Blastocyst 1 ICM feature ranked 17th. Second, the models disagreed markedly about which predictors matter (Fig.~\ref{fig::bymodel}). The tree ensembles concentrated importance on continuous physiological variables, chiefly female BMI, sperm DFI, female Gn-start-day progesterone, and male serum prolactin, while assigning almost no importance to the one-hot categorical indicators; logistic regression showed the opposite emphasis, ranking the categorical stimulation and laboratory indicators and the embryo-morphology features highest. This divergence is expected: axis-aligned trees split poorly on sparse binary dummies whereas a linear model can assign them large coefficients, so a single-model ranking would mislead. Third, the HAN attention weights spanned only about 0.50 to 0.56 across the 30 highest-ranked features, indicating mild global re-weighting rather than sharp feature selection; the ordering it induces should not be over-interpreted.

\begin{figure}[!t]
    \centering
    \IfFileExists{figures/fig_category_importance.pdf}{\includegraphics[width=\columnwidth]{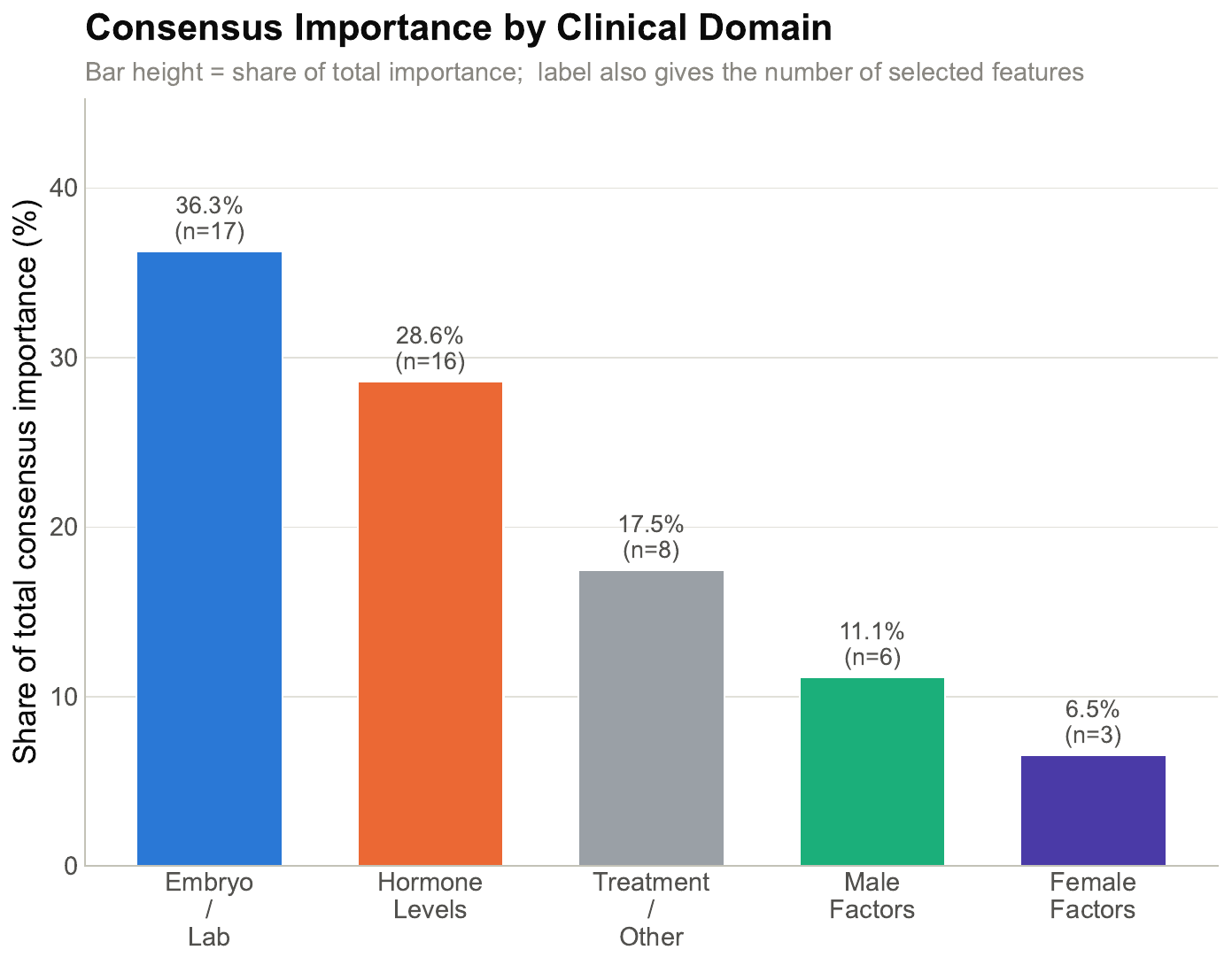}}{\fbox{\begin{minipage}[c][3.6cm][c]{0.86\columnwidth}\centering\textbf{Figure} --- add \texttt{figures/fig\_category\_importance.pdf}\par\smallskip\footnotesize category importance\end{minipage}}}
    \caption{Share of total consensus importance contributed by each of the four clinical domains defined in Table~\ref{tab::cohort}, with the number of selected features in each domain.}
    \label{fig::category}
\end{figure}

Aggregated by the same four clinical domains used in Table~\ref{tab::cohort}, oocyte-retrieval and embryo-development variables contributed the largest share of consensus importance, with the remainder distributed across female factors, stimulation and laboratory variables, and male factors (Fig.~\ref{fig::category}).

\subsection{Ablation}

Module-level ablations on the validation folds were consistent with the design choices. Removing the Feature Attention Module reduced validation AUC relative to the full backbone, supporting the contribution of the learned importance gate; the chosen dropout configuration (0.25 and 0.15) outperformed both no-dropout and heavier-dropout settings; and the grid search over mixing weights favoured a HAN-weighted combination ($w = 0.7$) over more balanced settings ($w = 0.5$ and $w = 0.3$ were lower on validation).

\subsection{Cross-Validation Stability}

The test-set ranking was consistent with the cross-validation used for model selection: the HAN + LR ensemble and the HAN backbone led under both, and the tree- and distance-based models trailed in both. This agreement, with three-seed averaging, supports the stability of the ordering, though absolute values carry wide confidence intervals at this cohort size.

\subsection{Language-Model Baselines}

Adapting pre-trained language models by text serialization was not competitive: the best reached an AUC of about 0.61, well below the classical and ensemble models, indicating that at this cohort size general-purpose text encoders do not capture the structured clinical signal as effectively as models operating on the tabular features directly.

\section{Discussion}\label{sec:discussion}

On a small, real-world IVF/ICSI cohort, a compact attention network combined with logistic regression predicted live birth more accurately than either component alone and than foundation-model and language-model baselines. The attention backbone was the stronger single component, the linear model contributed complementary structure, and a HAN-weighted average ($w = 0.7$) gave the best and most stable prediction. That combining logistic regression with the HAN backbone helped while combining it with TabPFN did not suggests the benefit comes from genuine diversity between the two models rather than from ensembling per se.

These findings match a broader literature reporting that, on small clinical tabular datasets, well-regularized classical models remain competitive with, and often superior to, elaborate deep architectures~\cite{tabpfnNat,benchmark}, and that careful feature engineering matters more than raw capacity. They also complement the predominantly image-based ART literature~\cite{mina,theilgaard,louis} by showing what is achievable from routinely captured structured data alone, aligning with work on advanced machine-learning paradigms for IVF prognostication~\cite{catalyzing} and with calls to weigh practical implementation considerations alongside raw accuracy~\cite{swain}.

The absolute discrimination achieved (AUC-ROC around 0.67 to 0.71) is in the range that is plausible, and arguably honest, for a strict live-birth endpoint predicted before transfer from a single-centre cohort. Studies built on large registries sometimes report higher figures~\cite{goyal,huang}, but they often predict easier endpoints, benefit from larger samples, or omit external validation; conversely, optimistic results in small studies frequently reflect leakage or the absence of a held-out test set. We aimed for a realistic rather than a maximal estimate of performance.

The choice of a hybrid was deliberate. A single large neural network overfits on a few hundred samples, while a purely linear model cannot capture interactions. Combining a compact attention network with logistic regression lets the convex combination act as a regularizer, which proved more effective than either component alone and than pairing the linear model with a foundation model. For small clinical cohorts, combining complementary simple models appears a more dependable route to robust performance than scaling up a single architecture.

The differential ICM/TE weighting is, to our knowledge, a novel encoding choice for this prediction task. Conceptually it resolves the tie between discordant blastocysts (for example 4AB versus 4BA) that equal-weight Gardner scoring cannot, and the TE-dominant weighting we found is consistent with evidence that trophectoderm quality is a strong correlate of implantation and live birth. We stress, however, that the association of the composite score with live birth was weak in absolute terms and that the weights were tuned on a single cohort; the scheme should therefore be regarded as a transparent, promising encoding to be validated externally rather than as a settled clinical scoring rule.

Within these limits, the study contributes a transparently described dataset and a reproducible model competitive with modern baselines while small enough (under 5,000 parameters) to train and audit easily~\cite{tripod,probast}.

The compactness of the model is itself a practical advantage: the ensemble trains in minutes on a single workstation and needs no specialized hardware at inference, lowering the barrier to local re-training and audit as regulators move toward lifecycle oversight of AI in reproductive medicine~\cite{medenica,guell}.

Before any clinical use, two steps are essential beyond discrimination. First, the probabilities should be formally calibrated, since a well-discriminating model can still be poorly calibrated and calibrated probabilities are what a counselling clinician needs. Second, the decision threshold should reflect the clinical costs of false positives and negatives rather than defaulting to 0.5. Both, with prospective evaluation, form part of the planned external-validation work.

\subsection{Clinical Interpretation of the Leading Predictors}

The predictors carrying the most weight are clinically coherent, supporting the model's face validity even though the analysis is associative. The leading role of the weighted trophectoderm grade agrees with evidence that trophectoderm quality tracks implantation and live birth at least as closely as the inner cell mass, and is the empirical counterpart of the encoding decision described above. The prominence of Gn-start-day progesterone matches the established observation that premature progesterone elevation compromises endometrial receptivity and depresses fresh-transfer success, motivating freeze-all strategies in affected cycles. Female age and body-mass index are firmly established determinants of oocyte competence, and the counts of punctured follicles, retrieved oocytes, and blastocysts index ovarian response and the size of the embryo cohort available for selection. Notably, two male factors, sperm DFI and serum prolactin, ranked highly in the tree ensembles, echoing a growing literature that paternal factors contribute measurably to live birth yet are under-represented in prediction models.

Two cautions attach. First, several of these variables are pathway-dependent: embryo counts and morphology are products of earlier steps in the same cycle, so their predictive value does not imply that intervening on them would change the outcome. Second, the disagreement between model families means no single ranking is definitive; we report the consensus and per-model detail so this uncertainty stays visible.

\section{Limitations}

Several limitations should be acknowledged. First, the dataset is single-centre and of modest size, which bounds generalizability and inflates the variance of subgroup estimates; external, multi-centre validation is required before any clinical use. Second, the design is retrospective. Third, the endpoint is binary live birth, which does not capture the full outcome pathway (e.g. biochemical pregnancy, clinical pregnancy, miscarriage). Fourth, the language-model baselines were evaluated with a single serialization scheme and may improve with more elaborate prompting or fine-tuning. Finally, our reliance on univariate feature selection and a single centre means the selected feature set may not transfer unchanged to other populations, and the importance analysis is associative and model-dependent rather than causal.

\section{Conclusion}

We developed a curated single-centre IVF/ICSI dataset, a differential ICM/TE encoding of blastocyst morphology that separates discordant grades which conventional equal-weight Gardner scoring cannot, and a HAN + LR ensemble for live-birth prediction after fresh embryo transfer. The ensemble achieved the best discrimination among eleven models, exceeding both its own components and contemporary tabular foundation-model and language-model baselines, while remaining compact and reproducible. The dataset, the encoding scheme, and the model provide a foundation for the next step, prospective, multi-centre external validation, which is necessary before such a tool can responsibly inform patient care.

\section*{Funding}
This research received no specific grant from any funding agency in the public, commercial, or not-for-profit sectors.

\section*{Data and Code Availability}
\textbf{Data.} The clinical dataset analysed in this study is \emph{not} publicly available. It consists of retrospective patient records from the Reproductive Medicine Center of Zhongnan Hospital, Wuhan University, and its release is restricted by patient-confidentiality requirements and by the conditions of the institutional ethics approval. A description of the dataset, including the variable list, the encoding scheme, and the conditions for access, is provided at \url{https://github.com/abdul-raheem-shahzad/ivf-live-birth-prediction/tree/main/data}. De-identified data may be made available to qualified researchers by the corresponding authors on reasonable request, subject to approval by the Ethics Committee of Zhongnan Hospital of Wuhan University and to a data-use agreement.

\textbf{Code.} All code required to reproduce the preprocessing, the embryo-morphology encoding, the models, and every table and figure in this paper is openly available at \url{https://github.com/abdul-raheem-shahzad/ivf-live-birth-prediction}.

\section*{Ethics Approval and Consent to Participate}
This study was approved by the Ethics Committee of Zhongnan Hospital of Wuhan University (Ethical Approval No.~2025346K). Due to the retrospective nature of the study, the requirement for informed consent was waived by the Ethics Committee.

\section*{Conflict of Interest}
The authors report no conflicts of interest.

\bibliographystyle{IEEEtran}
\bibliography{citation}

\end{document}